\documentclass{article} % For LaTeX2e
\usepackage[dvipsnames]{xcolor}
\usepackage[final]{neurips_2019}

\usepackage{amsmath,amsfonts,bm}

\def\eqref#1{equation~\ref{#1}}
\def\1{\bm{1}}

\DeclareMathAlphabet{\mathsfit}{\encodingdefault}{\sfdefault}{m}{sl}
\SetMathAlphabet{\mathsfit}{bold}{\encodingdefault}{\sfdefault}{bx}{n}

\usepackage{verbatim}
\usepackage{hyperref}
\usepackage{url}
\usepackage{graphicx}
\usepackage{wrapfig}
\usepackage{amssymb}
\usepackage{mathtools}
\usepackage{algorithm}
\usepackage[noend]{algpseudocode}
\usepackage{amsmath}
\usepackage{mathtools,nccmath}

\usepackage{algpseudocode,algorithmicx}

\newcommand{\cut}[1]{}
\makeatletter

\newcommand{\mypm}{\mathbin{\mathpalette\@mypm\relax}}
\newcommand{\@mypm}[2]{\ooalign{%
  \raisebox{.1\height}{$#1+$}\cr
  \smash{\raisebox{-.6\height}{$#1-$}}\cr}}
\makeatother
\title{Learning to Learn via Self-Critique}

\author{Antreas Antoniou \\
University of Edinburgh\\
\texttt{\{a.antoniou\}@sms.ed.ac.uk} \\
\And
Amos Storkey \\
University of Edinburgh\\
\texttt{\{a.storkey\}@ed.ac.uk}}
\newif\ifincludecomment
\includecommenttrue % \includecommenttrue or \includecommentflase
\ifincludecomment
  
\else
  \NewEnviron{guidance}{}{}
\fi

\usepackage[colorinlistoftodos,textsize=tiny, textwidth=18mm]{todonotes}
\ifincludecomment
\newcommand{\maybecomment}[1]{\todo[color=olive!40]{#1}} 
\newcommand{\maybetohere}[1]{\todo[color=red!40]{#1}} 
\newcommand{\maybedelete}[1]{\todo[color=blue!40]{#1}} 

\else
  \newcommand{\maybecomment}[1]{}
  
\newcommand{\maybedelete}[1]{} 
\fi
\newcommand{\amostohere}[1]{{\color{black}\maybetohere{AMOS HERE}}}

\newcommand{\proposedmethod}{SCA}
\newcommand{\proposedmethodspace}{SCA\ }

\begin{document}

\maketitle

\begin{abstract}
In few-shot learning, a machine learning system learns from a small set of labelled examples relating to a specific task, such that it can generalize to new examples of the same task. Given the limited availability of labelled examples in such tasks, we wish to make use of all the information we can. Usually a model learns task-specific information from a small training-set (\emph{support-set}) to predict on an unlabelled validation set (\emph{target-set}). The target-set contains additional task-specific information which is not utilized by existing few-shot learning methods. Making use of the target-set examples via transductive learning requires approaches beyond the current methods; at inference time, the target-set contains only unlabelled input data-points, and so discriminative learning cannot be used. In this paper, we propose a framework called \emph{Self-Critique and Adapt} or \proposedmethod, which learns to learn an label-free loss function, parameterized as a neural network. A base-model learns on a support-set using existing methods (e.g. stochastic gradient descent combined with the cross-entropy loss), and then is updated for the incoming target-task using the learnt loss function. The label-free loss function is learned such that the target-set-updated model achieves higher generalization performance. Experiments demonstrate that \proposedmethod\ offers substantially reduced error-rates compared to baselines which only adapt on the support-set, and results in state of the art benchmark performance on Mini-ImageNet and Caltech-UCSD Birds 200.

% For this reason we propose the use of transductive meta-learning for few-shot settings. Caltech-UCSD Birds 200 (CUB)
% Recent attempts towards solving the problem of few-shot learning, where only a handful of data is available to learn from, have been mainly focused on learning how a model can best learn task-specific information from a small training-set (often referred to as a \emph{support-set}), such that the resulting model can generalize very well on a small validation-set (also known as a \emph{target-set}). However, the target-set input data-points, for which predictions must be produced, contain additional task-specific information which are never utilized by existing methods. At inference time, the target-set contains only input data-points hence making effective learning on the target-set very hard with standard deep learning methods. In this paper, we propose a framework that can learn to learn a label-free loss function, parameterized as a neural network, which can be used to leverage target-set information, called \emph{Self-Critique and Adapt} or \proposedmethod. In effect, this allows a base-model to learn on a support-set using existing methods (i.e. stochastic gradient descent combined with the cross-entropy loss), and then update itself for the incoming target task using the learned loss function (i.e. the meta-learned label-free loss), such that it can achieve higher generalization performance. We carry experiments that demonstrate that our method offers higher generalization performance to our baselines which can only adapt on the support-set. 
\end{abstract}

\section{Introduction}
Humans can learn from a few data-points and generalize very well, but also have the ability to adapt in real-time to information from an incoming task. Given two training images for a two class problem, one with a cat on a white sofa, and one with a dog next to a car, one can come up with two hypothesis as to what each class describes in each case. A test image that presents a dog with a cat would be ambiguous. However, having other unlabelled test images with cats in other contexts and cars in other contexts enables these cases to be disambiguated, and it is possible to learn to focus on features that help this separation. We wish to incorporate this ability to adapt into a meta-learning context.

Few-shot learning is a learning paradigm where only a handful of samples are available to learn from. It is a setting where deep learning methods previously demonstrated weak generalization performance. In recent years, by framing the problem of few-shot learning as a meta-learning problem \citep{vinyals2016matching}, we have observed an advent of meta-learning methods that have demonstrated 
% \amos{For the figure and earlier, there are the aforementioned edits, but please also include a notation for the final loss}
unprecedented performance on a number of few-shot learning benchmarks 
\citep{snell2017prototypical, finn2017model, rusu2018meta}.

\begin{figure}[!htbp]
    \centering
    \includegraphics[width=1.0\textwidth]{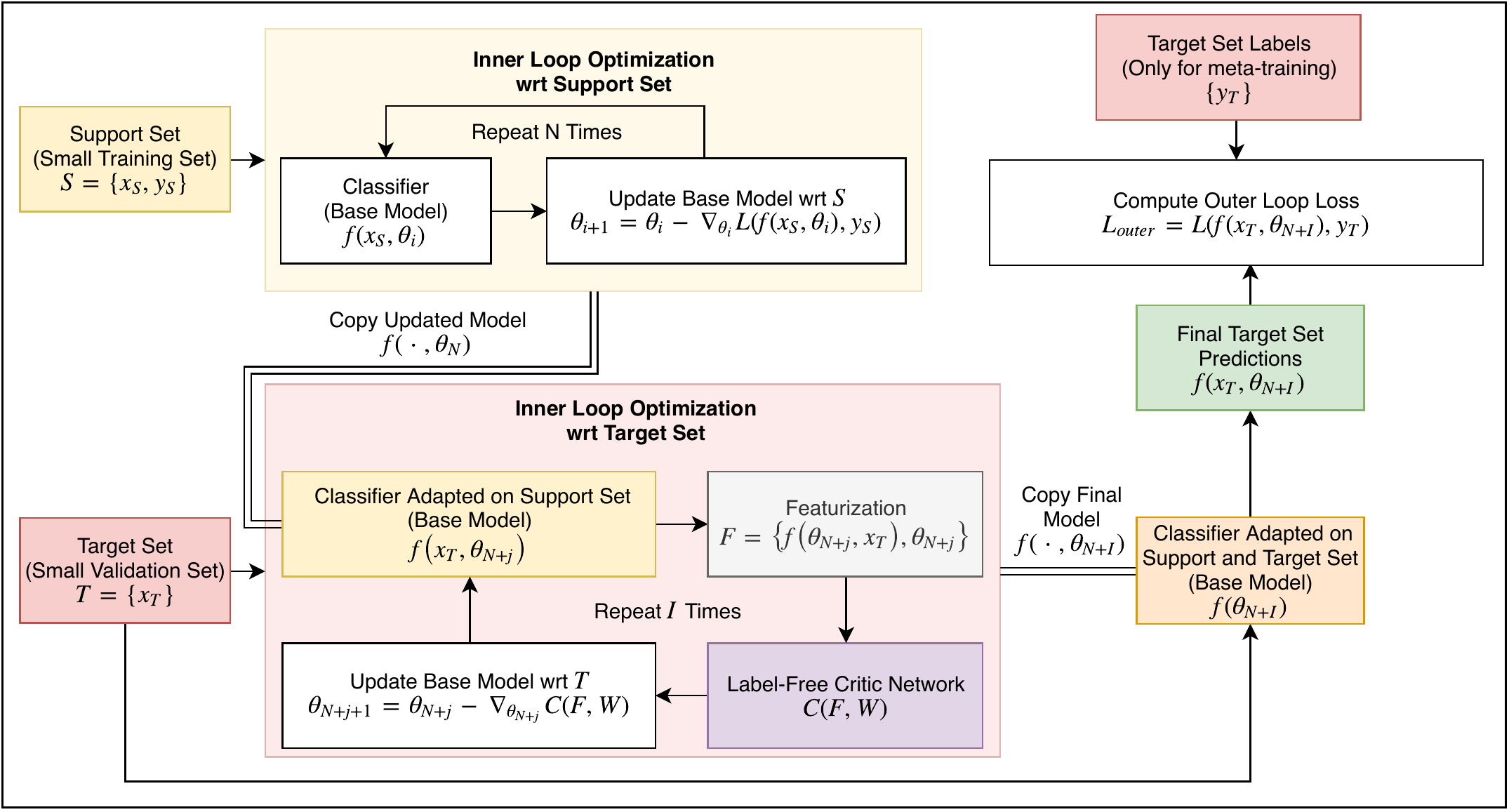}
    \vspace{-5.0mm}
    \caption{Proposed Method. Starting from the top-left, task-specific knowledge from the support-set is used to train the base model, updating $\theta_0$ to $\theta_N$. At this point, standard meta-learning methods return predictions from this learnt model to complete their inference. Instead, we use an unsupervised loss from a critic network $C$ applied to the unlabelled target set to make further updates. To do this we collect a set of features $F$ that summarise the model applied to the target set $T$; these features are sent to the critic $C$, a neural network with parameters $W$. Using the loss from this critic network, and the model with parameters $\theta_N$, we make further updates to get $\theta_{N+I}$. We use the predictions from this model as our predictions for the target set. During training, an \emph{outer-loop} loss comparing target-set predictions to target set labels is used to update the initial parameters $\theta_0$ and the critic model parameters $W$.}
    \label{fig:sca-diagram}
    \vspace{-5.0mm}
\end{figure}

Most few-shot meta-learning methods have focused on either learning static \citep{finn2017model,antoniou2018train,li2016learning} or dynamic parameter initializations \citep{rusu2018meta}, learning rate schedulers \citep{antoniou2018train}, embedding functions \cite{vinyals2016matching, snell2017prototypical}, optimizers \cite{ravi2016optimization} and other internal components. However all of these methods explore learning from the labelled support-set, whereas learning (at inference time) from the unlabelled target-set has remained unexplored.\footnote{Existing techniques like MAML \cite{finn2017model} utilize target-set information by computing means and standard deviations for the batch normalization layers within their base models. However, we don't consider that as explicit learning, but instead, as a minimal adaptation routine.}

In this paper, we propose a mechanism we call \emph{Self-Critique and Adapt} or \proposedmethod\ that enables meta-learning-based few-shot systems to learn not only from the support-set input-output pairs, but also from the target-set inputs, by learning a label-free loss function, parameterized as a neural network. Doing so grants our models the ability to learn from the target-set input data-points, simply by computing a loss, conditioned on base-model predictions of the target-set. The label-free loss can be used to compute gradients with respect to the model, and the gradients can then be used to update the base-model at inference time, to improve generalization performance. Furthermore, the proposed method can be added on top of any modern meta-learning method, including both methods that utilize gradient updates on the support set, such as MAML \citep{finn2017model}, as well as ones that do not use gradient-based updates on the support set, such as Matching Networks \citep{vinyals2016matching}.

Self-Critique and Adapt is a \emph{transductive learning} approach \citep{vapnik200624}. Transductive learning uses training data and test input data-points to learn a model that is \textbf{specifically} tuned to produce predictions for the given test-set. Transductive learning benefits from \emph{unsupervised information} from the test example points, and \emph{specification} by knowing where we need to focus model capability.
In stark contrast, \emph{inductive learning}, can be defined as a learning paradigm where given training input-output pairs, a model is learned consisting of \textbf{general} rules, that can then be used on any test-set without refinement to produce predictions. Given that, in a meta-learning context, additional learning needs to be done for each new setting anyway, and given the importance of making the most of every piece of information, transductive learning is a natural learning paradigm for the few-shot learning setting.

We evaluate the proposed method on the established few-shot learning benchmarks of Mini-ImageNet \citep{ravi2016optimization} and Caltech-UCSD Birds 200 (CUB) \citep{chen2019closerfewshot}. The evaluation results indicate that our method substantially boosts the performance of two separate instances of the MAML++ \citep{antoniou2018train} framework, setting a new state-of-the-art performance for all tasks in both benchmarks.

This paper's contributions are:

% learning a label-free loss function to leverage target-set information in order to improve few-shot learning performance
% conduct ablation studies to investigate which types of conditioning information produce the best performing critic networks

\begin{enumerate}
    \item An approach that gives state-of-the-art performance in the Mini-Imagenet and Caltech-UCSD Birds 200 (CUB) benchmark tasks by using both support and target set information through a transductive approach.
    \item The ability to learn to learn a flexible parameterized loss function appropriate for a supervised problem but defined on unlabelled data; this loss function can be used to enhance training on semi-supervised data.
    \item A set of ablation studies on different conditioning features for the critic network, revealing which features are most useful to the few-shot learning benchmarks.
\end{enumerate}

\section{Related Work}
% \amos{This should broaden beyond meta-learning approaches, and include a citation to all of our other few-shot learning stuff, including Neural Statistician.}
The \emph{set-to-set} few-shot learning setting \citep{vinyals2016matching} has been vital in framing few-shot learning as a meta-learning problem. In set-to-set few-shot learning, we have a number of tasks, which a model seeks to learn. Each task is composed of two small sets. A small training-set or \emph{support-set} used for acquiring task-specific knowledge, and a small validation-set or \emph{target-set}, which is used to evaluate the model once it has acquired knowledge from the support-set. The tasks are generated dynamically from a larger dataset of input-output pairs. The dataset is split into 3 subsets beforehand, the \emph{meta-training}, \emph{meta-validation} and the \emph{meta-test} sets, used for training, validation and testing respectively.

Once meta-learning was shown to be very effective in learning to few-shot learn, a multitude of methods showcasing unprecedented performance in few-shot learning surfaced. Matching Networks \citep{vinyals2016matching}, and their extension Prototypical Networks \citep{snell2017prototypical} were some of the first methods to showcase strong few-shot performance, using learnable embedding functions parameterized as neural networks in combination with distance metrics, such as cosine and euclidean distance. Unsupervised and supervised set-based embedding methods were also developed for the few-shot setting \citep{edwards2016towards}.

Further advancements were made by gradient-based meta-learning methods that explicitly optimized themselves for fast adaptation with a few data-points. The first of such methods, Meta-Learner LSTM \citep{ravi2016optimization} attempted to learn a parameter-initialization and an optimizer, parameterized as neural networks for fast adaptation. Subsequently, additional improvements came from \emph{Model Agnostic Meta-Learning} (MAML) \citep{finn2017model} and its improved versions \emph{Meta-SGD} \citep{li2017meta} and \emph{MAML++} \citep{antoniou2018train}, where the authors proposed learning a parameter-initialization for a base-model that is adapted with standard SGD for a number of steps towards a task. 

Furthermore, Relational Networks \citep{santoro2017simple}, that were originally invented for relational reasoning, demonstrated very strong performance in few-shot learning tasks \citep{santurkar2018does}. The effectiveness of relational networks in a large variety of settings made them a module often found in meta-learning systems.

In the Reinforcement Learning domain, meta-learning has been used to learn an unsupervised loss function for sample-efficient adaptation in \citep{yu2018towards}. Their work differs from ours, in that our model adapts a meta-learned initialization by both labelled and unlabelled examples; we do use a learnt unsupervised model, but the combination is critical. Their work is purely unsupervised in the inner loop, using RL as the outer loop, whereas ours consists of both supervised and unsupervised inner loops and a supervised outer loop. Furthermore, in \citep{houthooft2018evolved}, the authors propose a method that learns a state and reward conditional loss function to train a policy network, using RL inner phase and an evolutionary algorithm outer loop. Our work, instead targets a different problem, that is few-shot learning, using a supervised outer loop, and a transductive inner loop (composed by supervised and unsupervised inner loop phases). Furthermore, in \citep{floodcritic2017} the authors propose learning a supervised loss function for few-shot learning, as well as a state and reward conditional loss function for training RL agents.

% Perhaps add other papers too
Shortly after, substantial progress was made using a hybrid method utilizing embeddings, gradient-based methods and dynamic parameter generation called Latent Embedding Optimization \cite{rusu2018meta}. 

Semi-supervised learning via learning label-free functions were also attempted in \cite{semifewmaml2018}. Transductive learning for the few-shot learning setting has previously been attempted by learning to propagate labels \citep{liu2018transductive}. Their work differs from ours in that we learn to transduce using a learned unsupervised loss function whereas they instead learn to generate labels for the test set.
% \begin{fleqn}[\dimexpr\leftmargini-\labelsep]
%         \setlength\belowdisplayskip{0pt}
%         \begin{equation}
%             \begin{multlined}[c]
%               a = 222222222 + 222222222222222 + 222222 \\
%                 + 222222222222222222
%             \end{multlined}
%         \end{equation}
%         \end{fleqn}%
% \newenvironment{customequation}{\vspace*{-\baselineskip}\fleqn\setlength\belowdisplayskip{0pt}\equation\multilined}{\endmultilined\endequation\endfleqn}

\begin{algorithm}
\caption{SCA Algorithm combined with MAML \label{alg:full}}
\begin{algorithmic}[1]
\State $\mathbf{Required:}$ Base model function $\mathbf{f}$ and initialisation parameters $\boldsymbol{\theta}$, critic network function  $\mathbf{C}$ and parameters $\mathbf{W}$, a batch of tasks $\{\mathbf{S^{B}} = \{x_{S}^{B}, y_{S}^{B}\}, \mathbf{T^{B}} = \{x_{T}^{B}, y_{T}^{B}\}\}$ (where $\mathbf{B}$ is the number of tasks in our batch) and learning rates $\mathbf{\alpha, \beta, \gamma}$
% where \theta_{0} are the base initialization parameters and W are the critic's static parameters 
\State $L_{outer} = 0$
\For{b in range(B)}

\State $\theta_0 = \theta$ \Comment{Reset $\theta_0$ to the learned initialization parameters}
\For{i in range(N)} \Comment{N indicates total number of inner loop steps wrt support set}
\State \vspace*{-\baselineskip} \begin{fleqn}[\dimexpr\leftmargini-\labelsep] 
        \setlength\belowdisplayskip{0pt}
        \begin{gather} \label{equation:inner_support_update}
            \begin{multlined}[c]
                \theta_{i+1} = \theta_{i} - \alpha \nabla_{\theta_{i}} L(f(x_S^b, \theta_{i}), y_S^b)
            \end{multlined}
        \end{gather}
\end{fleqn}%
\Comment{Inner loop optimization wrt support set}
\EndFor

\For{j in range(I)}\Comment{I indicates total number of inner loop steps wrt target set}
\State \vspace*{-\baselineskip} \begin{fleqn}[\dimexpr\leftmargini-\labelsep] 
        \setlength\belowdisplayskip{0pt}
        \begin{gather} \label{equation:critic_features}
            \begin{multlined}[c]
                F = \{f(x_T^b, \theta_{N+j}), \theta_N+j, g(x_S, x_n)\}
            \end{multlined}
        \end{gather}
        \end{fleqn}%
\Comment{Critic feature-set collection}
\State \vspace*{-\baselineskip} \begin{fleqn}[\dimexpr\leftmargini-\labelsep] 
        \setlength\belowdisplayskip{0pt}
        \begin{gather} \label{equation:inner_target_update}
            \begin{multlined}[c]
                \theta_{N+j+1} = \theta_{N+j} - \gamma \nabla_{\theta_{N+j}} C(F, W)
            \end{multlined}
        \end{gather}
\end{fleqn}%
\Comment{Inner loop optimization wrt target set}
\EndFor

\State $L_{outer} = L_{outer} + L(f(x_{T}^b, \theta_{N+I}), y_{T}^b)$\label{maml_outer_objective}

\EndFor
\State \vspace*{-\baselineskip} \begin{fleqn} 
        \setlength\belowdisplayskip{0pt}
        \begin{gather} \label{maml_outer_update_base}
            \begin{multlined}
                \theta = \theta - \beta \nabla_{{\theta}} L_{outer}
            \end{multlined}
        \end{gather}
\end{fleqn}%
\Comment{Joint outer loop optimization of $\theta$}
\State \vspace*{-\baselineskip} \begin{fleqn}
        \setlength\belowdisplayskip{0pt}
        \begin{gather}  \label{maml_outer_update_critic}
            \begin{multlined}
                W = W - \beta \nabla_{{W}} L_{outer}
            \end{multlined}
        \end{gather}
\end{fleqn}%
\Comment{Joint outer loop optimization of $W$}
\end{algorithmic}
\end{algorithm}

\section{Self-Critique and Adapt}\label{section:model}
% \amos{Make Section headings more informative. E.g. Call the section by the name of your model}
For a model to learn and adapt in a setting where only input data-points are available (e.g. on given task's few-shot target-set), one needs a label-free loss function. For example, many unsupervised learning approaches try to maximize the generative probability, and hence use a negative log-likelihood (or bound thereof) as a loss function. In general though, much of the generative model will be task-irrelevant. In the context of a particular set of tasks there are likely to be more appropriate, specialised choices for a loss function.

Manually inventing such a loss function is challenging, often only yielding loss functions that might work in one setting but not in another. Understanding the full implications of a choice of loss-function is not easy. Instead, we propose a Self-Critique and Adapt approach which \emph{meta-learns} a loss function for a particular set of tasks. It does this by framing the problem using the set-to-set few-shot learning framework and using end-to-end differentiable gradient-based meta-learning as our learning framework.

\proposedmethod\ is model-agnostic, and can be applied on top of \textbf{any} end-to-end differentiable, gradient-based, meta-learning method that uses the inner-loop optimization process to acquire task-specific information. Many such approaches \citep{ravi2016optimization, finn2017model, li2017meta, antoniou2018train, finn2018probabilistic, qiao2018few, rusu2018meta, grant2018recasting} are currently competing for state-of-the-art in the few-shot learning landscape.

Self-Critique and Adapt, summarised in Figure~\ref{fig:sca-diagram}, takes a base-model, updates it with respect to the support-set with an existing gradient-based meta-learning method (e.g.  MAML \citep{finn2017model}, MAML++ \citep{antoniou2018train} or LEO \citep{rusu2018meta}), and then infers predictions for the target-set. Once the predictions have been inferred, they are concatenated along with other base-model related information (e.g. model parameters, a task embedding etc.), and are then passed to a learnable \emph{critic loss network}, the output of which should be interpreted as a loss value for the given inputs. This critic network computes and returns a loss with respect to the target-set. The base-model is then updated with any stochastic gradient optimization method (such as SGD) with respect to this critic loss; updates can be done a number of times if necessary. This \emph{inner-loop} optimization produces a predictive model specific to the support/target set information. 

The inner loop process is used directly at inference time for the task at hand. However, as in other meta-learning settings, we optimize the inner loop using a collection of training tasks (these are different from the test tasks as explained in Section \ref{section:experiments}).
The quality of the inner-loop learned predictive model is assessed using ground truth labels from the training tasks. The \emph{outer loop} then optimizes the initial parameters and the critic loss to maximize the quality of the inner loop predictions. As with other meta-learning methods, the differentiability of the whole inner loop ensures this outer-loop can be learnt with gradient based methods.

% \amos{Flag1}
% \amos{Perhaps the following bit is best as pseduocode? Or in an appendix. It is slightly unweildy at the moment.
% Seperate out the key concepts that need to be referred to elsewhere and the process that is implementational detail.
% }

% Reference to the Algorithm here. With basic description as to what it is. 
% Alg clean up.

In this paper we use MAML++ as the base method. We denote the model, which is parameterized as a neural network, by $f(\cdot,\theta)$, with parameters $\theta$, and the critic loss $C(\cdot,W)$, also a neural network with parameters $W$. We want to learn good parameters $\theta$ and $W$, such that when the model $f$ is optimized a number of steps $N$ with respect to the loss $L$ on the support-set $S_b=\{x_S, y_S\}$, and then additionally another $I$ steps towards the target-set $T_b=\{x_T\}$, using the critic loss $C$, it can achieve good generalization performance on the target-set. Here, $b$ is the index of a particular task in a batch of tasks.  The full algorithm is described in Algorithm~\ref{alg:full}.

Equation \ref{equation:critic_features} in Algorithm~\ref{alg:full} defines a potential set of conditioning features $F$ that summarise the base-model and its behaviour. These features are what the unsupervised critic loss $C$ can use to tune the target set updates. Amongst these possible features, $f(\theta_N, x_T)$ are the predictions of the base-model $f$, using parameters $\theta_N$ (that is parameters updated for N steps on the support-set loss) and $g(x_S, x_n)$ is a task embedding, parameterized as a neural network, which is conditional on the support and target input data-points.

\section{Example: A Prediction-Conditional Critic Network}
\label{sect:example}
\textbf{Model Inference and Training:}
 As explained in Sections \ref{section:model} and \ref{section:critic-features} our critic network can be given access to a wide variety of conditioning information. Here, we describe a specific variant, which uses the predictions of the base-model on the target-set as the critic network's conditioning information; this simple form enables visualisation of what the critic loss returns. 
 
 In this example, SCA is applied directly to MAML++. Given a support set, $S$ and target set $T$ for a task, and initial model $f(\cdot,\theta_0)$, five SGD steps are taken to obtain the updated model $f(\cdot,\theta_{5})$ using the support set's loss using Equation~\ref{equation:inner_support_update}. The updated model is then applied to the target set to produce target-set predictions $f(x_T,\theta_5)$; these predictions are passed to the critic network, which returns a loss-value for each element of the target set. Aggregated, these loss-values form the critic loss. The base model $f(\cdot,\theta_5)$ is then updated for 1 step using the gradients with respect to the critic loss using Equation \ref{equation:inner_target_update} to obtain the final model $f(\cdot,\theta_6)$. We use gradients of a whole batch of target set images to update the base model since this produces better performance. The results in the paper represent experiments where we used target sets of size 75 to learn the loss function, however one could instead randomly choose the batch size of the target set to train a loss function that generalizes on a wide range of batch sizes.
%  \amos{Could we distinguish between the initial model and the inner-loop optimized model. At the moment you call both the base-model, but they are different}. 
 At this stage our final model has learned from both the support and target sets and is ready to produce the final predictions of the target set. These are produced, and, at training time, evaluated against the ground truth. This final error is now the signal for updating the whole process: the choice of initial $\theta_0$ and the critic loss update. The gradients for both the base and critic model parameters can be computed using backpropagation as every module of the system is differentiable. 
 
\begin{figure}[!htbp]
     \centering
     \includegraphics[width=1.0\textwidth]{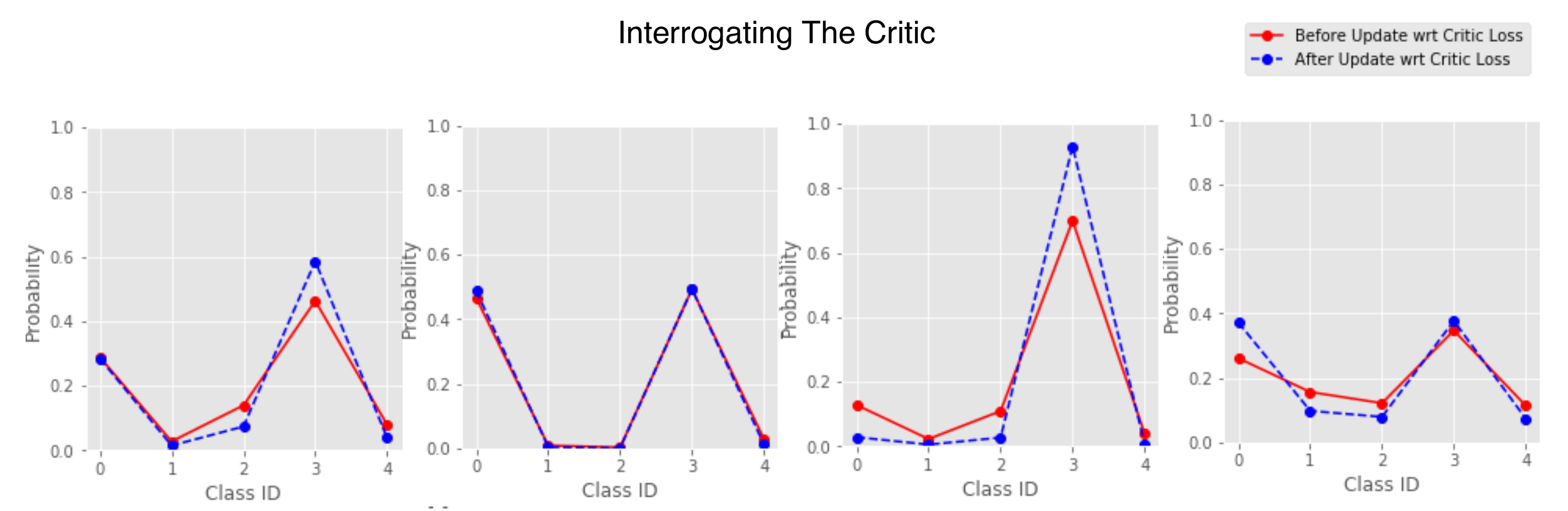}
     \caption{The target-set predictions of the base-model before (red lines) and after (green lines) it has been updated wrt the learned critic loss. Starting from the left, the first chart showcases an instance where the probabilities of two classes dominate the rest, with one class having approximately $20\%$ higher probability than the other. The critic network proposes that the dominant class with the lower probability be left as is, whilst the higher probability class be increased. In the second chart, we present an example where two dominant classes have equal probabilities, and are thus considered optimal and left unchanged. In the third case, we present an example where a single class dominates all others. The critic network proposes that the probability of that class is pushed to almost the maximum possible value. Finally, in the fourth case we present a very interesting case where the critic, faced with two dominant classes, which had approximately $10\%$ difference in their probability, proposed that the two classes have their probabilities matched to improve the resulting generalization performance. Note how some top classes change, e.g. in the rightmost chart. The effect of these changes is not in the class-labels alone but also allows better model initialization: training a critic on a pretrained initialization does not provide as much benefit. 
}
     \label{fig:critic_insight_mini}
     \vspace{-5.0mm}
 \end{figure}

\textbf{Interrogating the Learned Critic Loss:}
% A simple way to investigate the learned loss function would be to produce losses for a large variety of target-set predictions, and then manually inspect in an attempt to infer what the model considers as a strong or weak prediction. However doing so does not allow us to understand how the critic model would \textbf{modify} an existing prediction to improve it, which is what it actually does when applied at inference time.
We investigate how the predictions of the base-model change after updates with respect to the critic loss. We generate target-set predictions using the base-model $f(\cdot,\theta_5)$ and pass those predictions to the learnt critic network to obtain a loss value for each individual prediction. Five steps of SGD are performed with respect to the the individual critic loss. The results are shown in Figure \ref{fig:critic_insight_mini}.
% \amos{Ref figure}

% Figure \ref{fig:critic_insight_mini} illustrates three predictions optimized by the critic network. Each row indicates a particular target-set prediction optimized to minimize the critic's loss, starting from the original prediction in the first column with each column towards the right-hand side illustrating the prediction after an update step towards minimizing the critic's loss. Our critic in this instance has learned a prediction reqularizer. In the example illustrated in the third row we can see that our critic has clearly identified that one class in particular has been assigned a very high probability, and so it proposes to increase the probability of that class, while decreasing the others. In the second row, the critic has identified that two classes have been assigned approximately equal probabilities and thus the critic considers this prediction optimal, and proposes no changes. We theorize that changing anything in this particular instance had a very high risk of failure, hence why the model proposed no changes. Finally, in the first row, we can see an instance where there exist two classes's probabilities that appear to be dominating the others, however, contrary to the instance in the second row, in this instance one of the two classes has clearly significantly higher probability than the other. The critic proposes minimal increases in the top-most class, whilst keeping the second-highest class at its original probability. 

\raggedbottom
% three predictions optimized by the critic network. Each chart indicates a particular target-set prediction optimized to minimize the critic's loss, indicating the base-model's prediction prior to being optimized using the critic network in red, and after optimization in green. In the first chart, we can see an instance where the probabilities of two classes dominate the others with one class having significantly higher probability than the other. The critic proposes minimal increases in the top-most class, whilst keeping the second-highest class at its original probability. In the second chart, the critic has identified that two classes have been assigned approximately equal probabilities. We theorize that the critic considers modifying the probabilities to be too risky, hence proposing no changes. In the third row we can see that our critic has clearly identified that one class in particular has been assigned a very high probability, and so it proposes to increase the probability of that class, while decreasing the others.

% \amos{At this point you should give a very specific instantiation of your approach run it and show results.
% Then after that deal with the type of features you experimented with. You want the reader to understand a specific case of what you are doing as early in the document as possible.}

% \amos{Section title is unweildy: "features" seems to have 3 adjectives! Sections 4 and 5 are too verbose. Go for Simple and Concise. Say the absolute minimum you can get away with in this sort of section. Should Section 4 and 5 be the other way round?}

\section{Choosing Conditioning Information for the Critic Network}\label{section:critic-features}
We find that the performance of the critic network $C$, is unsurprisingly dependent on the quality of critic features it is given access to. In this section, we will outline a set of critic features we used in experiments, from simplest to most complicated; we also discuss the intuition behind their selection. To evaluate the usefulness of each set of critic features we use these in ablation studies in Section~\ref{section:experiments} and summarize the results in Tables~\ref{table:ablation} and~\ref{table:comparative}.

\textbf{Base-Model Predictions:}
The predictions on the base model indicate the certainty about any input; this can be a powerful driver for an unsupervised loss. As in Section~\ref{sect:example}, given the model $f$, the support-set updated weights $\theta_N$ and some target-set data-points $x_T$, we can generate predictions $f(x_T, \theta_N)$. These predictions can be passed to our critic model $C$ to compute the loss function.

\textbf{Base-model Parameters:}
Another key source of information about our base-model is the model parameters. In this context, the inner loop optimized parameters $\theta_N$ are passed to our loss network. For example this might enables it to learn \emph{penalty} terms, similar to manually invented penalties such as \emph{L1} and \emph{L2}.

\textbf{Task Embeddings:}\label{task-embedding-conditioning}
Giving the critic model direct access to task information can enable model assessment in the context of a particular task; we find this further improves performance as empirically observed in Tables \ref{table:ablation} and \ref{table:comparative}. To generate a task embedding, we learn an embedding function $g$, parameterized as a neural network such as a DenseNet.\footnote{Here we use a DenseNet, with growth rate 8, and 8 blocks per stage, for a total of 49 layers. The DenseNet is reqularised using dropout after each block (with drop probability of 0.5) and weight decay (with decay rate 2e-05).} The embedding function $g$ receives the support-set images and returns a batch of embedding vectors. Taking the mean of the vectors can serve as a strong task embedding. However, we found that using a relational network to relate all embeddings with each other and then taking the sum of those embeddings produced superior results. Similar embedding functions have previously been used in \citep{rusu2018meta}.

\section{Baselines}
The proposed model is, by virtue of its design, applicable to any few-shot meta-learning system that follows the set-to-set \cite{vinyals2016matching} few-shot learning setting. Thus, to investigate its performance, we require baseline meta-learning systems which inner-loop optimize their parameters only on the support-set. We chose the MAML++ system, for it's simplicity and strong generalization performance. 

To thoroughly evaluate our method, we experimented using two separate instances of the MAML++ framework, each differing only in the neural network architecture serving as its backbone. 

\textbf{Low-End MAML++:}
The backbone of our low-end baseline model follows the architecture proposed in \cite{antoniou2018train}, which consists of a cascade of 4 convolutional layers, each followed by a batch normalization layer and a ReLU activation function. We optimize the entirety of the backbone during the inner loop optimization process. The low-end baseline is chosen to be identical to an existing model (MAML++), such that we could: 1. Confirm that our implementation replicates the results of the original method (i.e. makes sure that our framework does not over/under perform relatively to an existing method, and thus reduces the probability that any improvements in our results are there due to a bug in the data provider or learning framework) and 2. Investigate how our proposed method performs when added adhoc to an existing, non-tuned, architecture, therefore showcasing performance unbiased wrt architecture.

\textbf{High-End MAML++:}
Methods that provide significant performance improvements on low-capacity models, often fail to provide the same level of improvement on high-capacity models and vice versa. Furthermore, meta-learning methods, have been demonstrated to be very sensitive to neural network architecture changes. Thus, to evaluate both the consistency and sensitivity of our method, we designed a high (generalization) performance MAML++ backbone. It uses a shallow, yet wide DenseNet architecture, with growth-rate 64, utilizing 2 dense-stages each consisting of two bottleneck blocks as described in \cite{huang2017densely} and one transition layer. Before each bottleneck block, we apply squeeze-excite style convolutional attention as described in \cite{hu2018squeeze}, to further regularize our model. To improve the training speed and generalization performance, we restrict the network components optimized during the inner loop optimization process. In more detail, we choose to share a static copy of majority of the network components at each step, and only optimize the penultimate convolutional layer (including the squeeze excite attentional block preceding it) as well as the final linear layer. An efficient way of sharing components across steps, is to treat them as a \emph{feature} embedding, whose features are passed to the components that will be inner loop optimized. Recent work \citep{rusu2018meta,qiao2018few} followed a similar approach. Motivations behind these design choices can be found in the supplementary materials section 1. \ref{appendix-high-end-design-choices}.

\textbf{Critic Network Architecture:}
The critic network consists of two majour components. First, a selection of conditioning features as described in Section \ref{section:critic-features}, which are then reshaped into a batch of one-dimensional features vectors and concatenated on the feature dimension. Second, an \emph{Information Integration} network, which consists of a sequence of five one-dimensional dilated convolutions with kernel size 2, and 8 kernels per convolutional layer. Further, we employ an exponentially increasing dilation policy where a given convolutional layer has dilation $d = 2^i$ where $i$ is the index of the convolutional layer in the network, starting from $0$ for the first layer and increasing up to $4$ for the fifth layer. We use Dense-Net style connectivity for our convolutional layers, more specifically, the inputs for a given layer consist of a concatenation of the output features of all preceding convolutional layers. Finally we apply a sequence of two fully connected layers using ReLU non-linearities before the final linear layer which outputs the loss value.
% in more detail, for the former architecture, we optimized the entirety of the architecture during the inner loop optimization, whereas for the densenet, we shared all layers across time-steps, with the exception of the penulimate (a convolutional) layer and final linear layer. We chose to do this, as the learned inner loop learning rates, clearly indicated that the only layers that had a high learning rate where the last and penultimate layers. Hence, optimizing only those, decreases the computational overheards, whilst keeping the generalization performance to a similar level. Furthermore, given recent insights from {}, we decided to use fewer dimensionality reduction layers, and overall shallower but wider densenets, that would learn features that are less class-specific and thus have a higher probability of generalizing well to unseen classes. Finally, the convolutional layer optimized in the inner loop optimization process also included a squeeze-excite attentional mechanism that we found improves generalization performance. Our second baseline, achieved results comparable to the current state of the art, and was suitable for evaluating our proposed model on systems that have already benefitted from a deeper backbone and more sophisticated architectures. It is worth noting that we originally intended to use the LEO meta-learning system as our second baseline, however, due to failure in replicating the original paper's results, we decided to use a simpler, yet just as robust MAML++ with a shallow and wide squeeze-excite densenet. would yield i 

\section{Experiments}\label{section:experiments}
To evaluate the proposed methods we first establish baselines on both the low-end and high-end variants of MAML++ on the Mini-ImageNet and Caltech-UCSD Birds 200 (CUB) 5-way 1/5-shot tasks. Then, we add the proposed critic network, which enables the adaptation of the base-model on a given target-set. To investigate how the selection of conditional information we provide to the critic affect its performance we conduct ablation studies. Due to the large number of combinations possible, we adopt a hierarchical experimentation style, where we first ran experiments with a single source of conditional information, and then tried combinations of the best performing methods. Finally, we ran experiments where we combined every proposed conditioning method to the critic, as well as experiments where we combined every conditional method except one. Tables \ref{table:ablation} and \ref{table:comparative} show ablation and comparative studies on Mini-Imagenet and CUB respectively.
% \amos{This is where experimental details should be. Particular choices etc.}
\subsection{Results}
The results of our experiments showcased that our proposed method was able to substantially improve the performance of both our baselines across all Mini-ImageNet and CUB tasks. 

More specifically, the original MAML++ architecture (i.e. Low-End MAML++), yielded superior performance when we provided additional conditional information such as the task embedding and the model parameters. The only source of information that actually decreased performance were the base-model's parameters themselves. Furthermore, it appears that a very straight-forward strategy that worked the best, was to simply combine all proposed conditioning sources and pass them to the critic network.

However, in the case of the High-End MAML++ architecture, performance improvements were substantial when using the predictions and the task embedding as the conditioning information. Contrary to the results on MAML++, providing the critic with the matching network and relational comparator features (in addition to the prediction and task embedding features) did not produce additional performance gains.

\begin{table}[!htb]
\centering
\scalebox{0.67}{%
\begin{tabular}{ l | c c c c} 
 \hline \\ [-1.5ex]
 \textbf{Model}                     & \multicolumn{4}{c}{\textbf{Test Accuracy}} \\  & \multicolumn{2}{c}{\textbf{Mini-Imagenet}} & \multicolumn{2}{c}{\textbf{CUB}} \\
  & 1-shot & 5-shot & 1-shot & 5-shot \\
 \hline \\ [-1.5ex]
 MAML++ (Low-End)                                                           & $52.15\mypm0.26\%$    & $68.32\mypm0.44\%$ & $62.19\mypm0.53\%$ & $76.08\mypm0.51\%$ \\ \hline
 MAML++ (Low-End) with \proposedmethod(preds)                               & $52.52\mypm1.13\%$    & $70.84\mypm0.34\%$ & $66.13\mypm0.97\%$ & $77.62\mypm0.77\%$ \\ 
 MAML++ (Low-End) with \proposedmethod(preds, params)                       & $52.68\mypm0.93\%$    & $69.83\mypm1.18\%$ & - & - \\  
 MAML++ (Low-End) with \proposedmethod(preds, task-embedding)               & $\mathbf{54.84\mypm1.24\%}$    & $70.95\mypm0.17\%$ & $65.56\mypm0.48\%$ & $77.69\mypm0.47\%$ \\
 MAML++ (Low-End) with \proposedmethod(preds, task-embedding, params)       & $54.24\mypm0.99\%$    & $\mathbf{71.85\mypm0.53\%}$ & - & - \\ \hline
 MAML++ (High-End)	                                                        & $58.37\mypm0.27\%$	& $75.50\mypm0.19\%$   & $67.48\mypm1.44\%$	  & $83.80\mypm0.35\%$ \\ \hline
 MAML++ (High-End) with \proposedmethodspace(preds)	                        & $\mathbf{62.86\mypm0.70\%}$	& $77.07\mypm0.19\%$ & $70.33\mypm0.78\%$    & $85.47\mypm0.40\%$            \\ 
 MAML++ (High-End) with \proposedmethodspace(preds, task-embedding)	        & $62.29\mypm0.38\%$	& $\mathbf{77.64\mypm0.40\%}$ & $\mathbf{70.46\mypm1.18\%}$    & $\mathbf{85.63\mypm0.66\%}$   \\
%  MAML++ (High-End) with \proposedmethodspace(preds, task-embedding, matching-network)	                                & $62.03\mypm0.38\%$	& $76.77\mypm0.40\%$ \\
%  MAML++ (High-End) with \proposedmethodspace(preds, task-embedding, matching-network, relational-comparator-network)	& $62.93\mypm0.19\%$    & $77.23\mypm0.54\%$ \\
 
 \hline
\end{tabular}
}
\vspace{0.5mm}
\caption{\proposedmethodspace Ablation Studies on Mini-ImageNet and CUB: All variants utilizing the proposed SCA method perform substantially better than the non-SCA baseline variant. Interestingly, the best type of critic conditioning features varies depending on the backbone architecture. Based on our experiments, the best critic conditioning features for the Low-End MAML++ is the combination of predictions, task-embedding and network parameters, whereas on High-End MAML++, using just the target-set predictions appears to be enough to obtain the highest performance observed in our experiments.}
\label{table:ablation}
\vspace{-4.0mm}
\end{table}

\begin{table}[!htb]
\centering
\scalebox{0.75}{%
\begin{tabular}{ l | c c c c} 
 \hline \\ [-1.5ex]
 \textbf{Model}                   & \multicolumn{4}{c}{\textbf{Test Accuracy}} \\ & \multicolumn{2}{c}{\textbf{Mini-ImageNet}} & \multicolumn{2}{c}{\textbf{CUB}} \\ 
  & 1-shot & 5-shot & 1-shot & 5-shot \\
 \hline \\ [-1.5ex]
 Matching networks~\citep{vinyals2016matching}  & $43.56 \mypm0.84\%$        & $55.31 \mypm0.73\%$ & $61.16 \mypm0.89\%$& $72.86 \mypm0.70\%$ \\
 Meta-learner LSTM~\citep{ravi2016optimization} & $43.44 \mypm0.77\%$        & $60.60 \mypm0.71\%$ &-&-\\
 MAML \citep{finn2017model}                     & $48.70 \mypm1.84\%$        & $63.11 \mypm0.92\%$ & $55.92 \mypm0.95\%$                 & $72.09 \mypm0.76\%$ \\
 LLAMA \citep{grant2018recasting}               & $49.40 \mypm1.83\%$        & - &-&-\\
 REPTILE \citep{nichol2018first}                & $49.97 \mypm 0.32\%$         & $65.99 \mypm 0.58\%$ &-&-\\
 PLATIPUS \citep{finn2018probabilistic}         & $50.13 \mypm 1.86\%$         & - &-&-\\
  \hline \\ [-1.5ex]
 Meta-SGD (our features)                        & $54.24 \mypm 0.03\%$ & $70.86 \mypm 0.04\%$ &-&-\\
 SNAIL~\citep{mishra2017simple}                 & $55.71 \mypm 0.99\%$ & $68.88 \mypm 0.92\%$ &-&-\\ 
 \citep{gidaris2018dynamic}                     & $56.20 \mypm 0.86\%$ & $73.00 \mypm 0.64\%$ &-&-\\ 
 \citep{munkhdalai2017meta}                     & $57.10 \mypm 0.70\%$ & $70.04 \mypm 0.63\%$ &-&-\\
 TADAM \citep{oreshkin2018tadam}                & $58.50 \mypm 0.30\%$ & $76.70 \mypm 0.30 \% $ &-&-\\
 \citep{qiao2018few}                            & $59.60 \mypm 0.41\%$ & $73.74 \mypm 0.19\%$ &-&-\\
 LEO \citep{rusu2018meta}                       & $61.76 \mypm 0.08\%$ & $77.59 \mypm 0.12\%$ &-&-\\ \hline
 Baseline \citep{chen2019closerfewshot}         &-&-                            & $47.12 \mypm0.74\%$   & $64.16 \mypm0.71\%$ \\
 Baseline ++ \citep{chen2019closerfewshot}      &-&-                            & $60.53 \mypm0.83\%$   & $79.34 \mypm0.61\%$ \\
 MAML (Local Replication)                       & $48.25 \mypm0.62\%$        & $64.39 \mypm0.31\%$ &-&-\\ 
 MAML++ (Low-End)                               & $52.15 \mypm0.26\%$        & $68.32 \mypm0.44\%$ & $62.19\mypm0.53\%$ & $76.08\mypm0.51\%$\\ 
 MAML++ (Low-End) + \proposedmethodspace        & $54.84 \mypm0.99\%$        & $71.85 \mypm0.53\%$ & $66.13\mypm0.97\%$ & $77.62\mypm0.77\%$\\ \hline
 MAML++ (High-End)                              & $58.37 \mypm0.27\%$        & $75.50 \mypm0.19\%$ & $67.48 \mypm1.44\%$	              & $83.80 \mypm0.35\%$ \\ 
 \textbf{MAML++ (High-End) + \proposedmethod}   & $\mathbf{62.86 \mypm0.79\%}$& $\mathbf{77.64 \mypm0.40\%}$ & $\mathbf{70.46 \mypm1.18\%}$        & $\mathbf{85.63 \mypm0.66\%}$ \\
 \hline
\end{tabular}
}
\vspace{1.0mm}
\caption{Comparative Results on Mini-ImageNet and CUB: The proposed method appears to improve the baseline model by over 4 percentage points, allowing it to set a new state-of-the-art result on both the 1/5-way Mini-ImageNet tasks.}
\label{table:comparative}
\vspace{-4.0mm}
\end{table}

\section{Conclusion}
In this paper we propose adapting few-shot models not only on a given support-set, but also on a given target-set by learning a label-free critic model. In our experiments we found that such a critic network can, in fact, improve the performance of two instances of the well established gradient-based meta-learning method MAML. We found that some of the most useful conditional information for the critic model were the base-model's predictions, a relational task embedding and a relational support-target-set network. The performance achieved by the High-End MAML++ with \proposedmethodspace is the current best across all SOTA results. The fact that a model can learn to update itself with respect to an incoming batch of data-points is an intriquing one. Perhaps deep learning models with the ability to adapt themselves in light of new data, might provide a future avenue to flexible, self-updating systems that can utilize incoming data-points to improve their own performance at a given task.

\section{Acknowledgements}
We would like to thank our colleagues Elliot Crowley, Paul Micaelli, James Owers and Joseph Mellor for reviewing this work and providing useful suggestions/comments. Furthermore, we'd like to thank Harri Edwards for the useful discussions at the beginning of this project and the review and comments he provided. This work was supported in part by the EPSRC Centre for Doctoral Training in Data Science, funded by the UK Engineering and Physical Sciences Research Council (grant EP/L016427/1) and the University of Edinburgh as well as a Huawei DDMPLab Innovation Research Grant.

\bibliographystyle{apalike}
\bibliography{iclr2019_conference,datasets,deeplearning,general,combined_metalearning}

\begin{thebibliography}{}

\bibitem[Antoniou et~al., 2019]{antoniou2018train}
Antoniou, A., Edwards, H., and Storkey, A. (2019).
\newblock How to train your {MAML}.
\newblock {\em In International Conference on Learning Representations}.

\bibitem[Chen et~al., 2019]{chen2019closerfewshot}
Chen, W.-Y., Liu, Y.-C., Kira, Z., Wang, Y.-C., and Huang, J.-B. (2019).
\newblock A closer look at few-shot classification.
\newblock In {\em International Conference on Learning Representations}.

\bibitem[Edwards and Storkey, 2017]{edwards2016towards}
Edwards, H. and Storkey, A. (2017).
\newblock Towards a neural statistician.
\newblock In {\em International Conference on Learning Representations (ICLR)
  (and arXiv:1606.02185 2016)}.

\bibitem[Finn et~al., 2017]{finn2017model}
Finn, C., Abbeel, P., and Levine, S. (2017).
\newblock Model-agnostic meta-learning for fast adaptation of deep networks.
\newblock {\em arXiv preprint arXiv:1703.03400}.

\bibitem[Finn et~al., 2018]{finn2018probabilistic}
Finn, C., Xu, K., and Levine, S. (2018).
\newblock Probabilistic model-agnostic meta-learning.
\newblock In {\em Advances in Neural Information Processing Systems}, pages
  9516--9527.

\bibitem[Gidaris and Komodakis, 2018]{gidaris2018dynamic}
Gidaris, S. and Komodakis, N. (2018).
\newblock Dynamic few-shot visual learning without forgetting.
\newblock In {\em Proceedings of the IEEE Conference on Computer Vision and
  Pattern Recognition}, pages 4367--4375.

\bibitem[Grant et~al., 2018]{grant2018recasting}
Grant, E., Finn, C., Levine, S., Darrell, T., and Griffiths, T. (2018).
\newblock Recasting gradient-based meta-learning as hierarchical bayes.
\newblock {\em arXiv preprint arXiv:1801.08930}.

\bibitem[Houthooft et~al., 2018]{houthooft2018evolved}
Houthooft, R., Chen, Y., Isola, P., Stadie, B., Wolski, F., Ho, O.~J., and
  Abbeel, P. (2018).
\newblock Evolved policy gradients.
\newblock In {\em Advances in Neural Information Processing Systems}, pages
  5400--5409.

\bibitem[Hu et~al., 2018]{hu2018squeeze}
Hu, J., Shen, L., and Sun, G. (2018).
\newblock Squeeze-and-excitation networks.
\newblock In {\em Proceedings of the IEEE conference on computer vision and
  pattern recognition}, pages 7132--7141.

\bibitem[Huang et~al., 2017]{huang2017densely}
Huang, G., Liu, Z., Van Der~Maaten, L., and Weinberger, K.~Q. (2017).
\newblock Densely connected convolutional networks.
\newblock In {\em CVPR}, volume~1, page~3.

\bibitem[Li and Malik, 2016]{li2016learning}
Li, K. and Malik, J. (2016).
\newblock Learning to optimize.
\newblock {\em arXiv preprint arXiv:1606.01885}.

\bibitem[Li et~al., 2017]{li2017meta}
Li, Z., Zhou, F., Chen, F., and Li, H. (2017).
\newblock Meta-sgd: Learning to learn quickly for few shot learning.
\newblock {\em arXiv preprint arXiv:1707.09835}.

\bibitem[Liu et~al., 2018]{liu2018transductive}
Liu, Y., Lee, J., Park, M., Kim, S., and Yang, Y. (2018).
\newblock Transductive propagation network for few-shot learning.
\newblock {\em arXiv preprint arXiv:1805.10002}.

\bibitem[Mishra et~al., 2017]{mishra2017simple}
Mishra, N., Rohaninejad, M., Chen, X., and Abbeel, P. (2017).
\newblock A simple neural attentive meta-learner.
\newblock {\em arXiv preprint arXiv:1707.03141}.

\bibitem[Munkhdalai and Yu, 2017]{munkhdalai2017meta}
Munkhdalai, T. and Yu, H. (2017).
\newblock Meta networks.
\newblock {\em arXiv preprint arXiv:1703.00837}.

\bibitem[Nichol et~al., 2018]{nichol2018first}
Nichol, A., Achiam, J., and Schulman, J. (2018).
\newblock On first-order meta-learning algorithms.
\newblock {\em arXiv preprint arXiv:1803.02999}.

\bibitem[Oreshkin et~al., 2018]{oreshkin2018tadam}
Oreshkin, B., L{\'o}pez, P.~R., and Lacoste, A. (2018).
\newblock Tadam: Task dependent adaptive metric for improved few-shot learning.
\newblock In {\em Advances in Neural Information Processing Systems}, pages
  719--729.

\bibitem[Qiao et~al., 2018]{qiao2018few}
Qiao, S., Liu, C., Shen, W., and Yuille, A.~L. (2018).
\newblock Few-shot image recognition by predicting parameters from activations.
\newblock In {\em Proceedings of the IEEE Conference on Computer Vision and
  Pattern Recognition}, pages 7229--7238.

\bibitem[Ravi and Larochelle, 2016]{ravi2016optimization}
Ravi, S. and Larochelle, H. (2016).
\newblock Optimization as a model for few-shot learning.

\bibitem[Rinu~Boney, 2018]{semifewmaml2018}
Rinu~Boney, A.~I. (2018).
\newblock Semi-supervised few-shot learning with maml.
\newblock {\em ICLR2018, Meta-Learning Workshop}.

\bibitem[Rusu et~al., 2018]{rusu2018meta}
Rusu, A.~A., Rao, D., Sygnowski, J., Vinyals, O., Pascanu, R., Osindero, S.,
  and Hadsell, R. (2018).
\newblock Meta-learning with latent embedding optimization.
\newblock {\em arXiv preprint arXiv:1807.05960}.

\bibitem[Santoro et~al., 2017]{santoro2017simple}
Santoro, A., Raposo, D., Barrett, D.~G., Malinowski, M., Pascanu, R.,
  Battaglia, P., and Lillicrap, T. (2017).
\newblock A simple neural network module for relational reasoning.
\newblock In {\em Advances in neural information processing systems}, pages
  4967--4976.

\bibitem[Santurkar et~al., 2018]{santurkar2018does}
Santurkar, S., Tsipras, D., Ilyas, A., and Madry, A. (2018).
\newblock How does batch normalization help optimization?(no, it is not about
  internal covariate shift).
\newblock {\em arXiv preprint arXiv:1805.11604}.

\bibitem[Snell et~al., 2017]{snell2017prototypical}
Snell, J., Swersky, K., and Zemel, R. (2017).
\newblock Prototypical networks for few-shot learning.
\newblock In {\em Advances in Neural Information Processing Systems}, pages
  4077--4087.

\bibitem[Sung et~al., 2017]{floodcritic2017}
Sung, F., Zhang, L., Xiang, T., Hospedales, T.~M., and Yang, Y. (2017).
\newblock Learning to learn: Meta-critic networks for sample efficient
  learning.
\newblock {\em CoRR}, abs/1706.09529.

\bibitem[Vapnik, 2006]{vapnik200624}
Vapnik, V. (2006).
\newblock 24 transductive inference and semi-supervised learning.

\bibitem[Vinyals et~al., 2016]{vinyals2016matching}
Vinyals, O., Blundell, C., Lillicrap, T., Wierstra, D., et~al. (2016).
\newblock Matching networks for one shot learning.
\newblock In {\em Advances in Neural Information Processing Systems}, pages
  3630--3638.

\bibitem[Yu, 2018]{yu2018towards}
Yu, Y. (2018).
\newblock Towards sample efficient reinforcement learning.

\end{thebibliography}
\bibliographystyle{apalike}
% \appendix

\section{Appendix: High-End Backbone details:}\label{appendix-high-end-design-choices}
The motivations behind each of the design choices can be found below.
\begin{enumerate}
    \item Using DenseNet as the backbone, which decreases probability of gradient degradation problems and by allowing feature-reuse across all blocksm improves parameter/training efficiency. MAML is highly vulnerable to gradient degradation issues, and thus ensuring that our backbone decreases probability of such issues is of vital importance.
    \item Using a shallow, yet wide backbone: Previous works \cite{qiao2018few,rusu2018meta} have demonstrated that using features from the 20th layer of a pretrained ResNet produces superior generalization performance. The authors made the case that using features from shallower parts of the network decreases the probability that the features are too class-specific, and thus allow for better generalization on previously unseen classes. In both \cite{qiao2018few,rusu2018meta} the authors did not train their meta-learning system end-to-end, and instead trained the feature backbone and the meta-learning components separately. However, in preliminary experiments we found that ResNet and DenseNet backbones tend to overfit very heavily, and in pursuit of a high-generalization end-to-end trainable meta-learning system, we experimented with explicit reduction of the effective input region of the layers in a backbone. Doing so, ensures that features learned will be local. We found that keeping the effective input region of the deepest layer to approximately 15x15/20x20 produced the best results for both Mini-ImageNet and CUB. Furthermore, we found that widening the network produced additional generalization improvements. We theorize that this is because of a higher probability for a randomly initialized feature to lie in just the right subspace to produce a highly generalizable feature once optimized.
    \item Using bottleneck blocks, preceded by squeeze-excite-style\cite{hu2018squeeze} convolutional attention: We empirically found that this improves generalization performance.
    \item Inner-Loop optimize only the last squeeze excite linear layer, as well as the last convolutional layer and the final linear layer, whilst sharing the rest of the backbone across the inner loop steps. This design choice was hinted by the learned per-layer, per-step learning rates learned by MAML++ on the low-end baseline. More specifically, the learned learning rates where close to zero, for all layers, in all steps, except the last convolutional and last linear layers. Thus, we attempted to train a MAML++ instance where only those two layers where optimized in the inner loop, while the rest of the layers where shared across steps. In doing so, we found that doing so makes no difference to generalization performance, whilst increasing the training and inference speeds by at least 15 times. 
\end{enumerate}
\end{document}